\documentclass[10pt,twocolumn,letterpaper]{article}
\usepackage{cvpr}
\usepackage{times}
\usepackage{epsfig}
\usepackage{graphicx}
\usepackage{amsmath}
\usepackage{amssymb}
\usepackage{multirow}
\usepackage{float}
\usepackage{tabulary}
\usepackage{tabularx}
\usepackage[tight,normalsize,sf,SF]{subfigure}
\usepackage{color}
\usepackage{soul}
\usepackage{enumitem}
\usepackage{hhline}
\usepackage{algorithm}
\usepackage{algpseudocode}
\usepackage{array}
\usepackage{flushend}
\usepackage{csquotes}
\newcolumntype{P}[1]{>{\centering\arraybackslash}p{#1}}
\DeclareMathSymbol{\Lambda}{\mathalpha}{operators}{3}
\DeclareMathSymbol{\Pi}{\mathalpha}{operators}{5}
\graphicspath{{./figs/}}
\DeclareGraphicsExtensions{.eps,.ps,.pdf,.png,.tiff}

\usepackage[table]{xcolor}
\usepackage[pagebackref=true,breaklinks=true,letterpaper=true,colorlinks,bookmarks=false]{hyperref}

 \cvprfinalcopy 
\def\cvprPaperID{3324}

\ifcvprfinal\fi
\begin{document}

\title{Collaborative Summarization of Topic-Related Videos}

\author{Rameswar Panda and Amit K. Roy-Chowdhury\\
Department of ECE, UC Riverside\\
{\tt\small rpand002@ucr.edu, amitrc@ece.ucr.edu}
}

\maketitle

\begin{abstract} 
Large collections of videos are grouped into clusters
by a topic keyword, such as  \enquote{Eiffel Tower} or \enquote{Surfing}, with many important visual concepts repeating across them. Such a topically close set of videos have mutual influence on each other, which could be used to summarize one of them by exploiting information from others in the set. We build on this intuition to develop a novel approach to extract a summary that simultaneously captures both important particularities arising in the given video, as well as, generalities identified from the set of videos. The topic-related videos provide visual context to identify the important parts of the video being summarized. We achieve this by developing a collaborative sparse optimization method which can be efficiently solved by a half-quadratic minimization algorithm. Our work builds upon the idea of collaborative techniques from information retrieval and natural language processing, which typically use the attributes of other similar objects to predict the attribute of a given object. Experiments on two challenging and diverse datasets well demonstrate the efficacy of our approach over state-of-the-art methods.

\end{abstract}
\vspace{-4mm}
\section{Introduction}
\label{sec:intro}
\vspace{-1mm}
With the recent explosion of \enquote{big (video) data} over the Internet, it is becoming increasingly important to automatically extract brief yet informative video summaries in order to enable a more efficient and engaging viewing experience. 
As a result, \textit{video summarization}, that automates this process, has attracted intense attention in the recent years.
\begin{figure}
	\centering
	\begin{tabular}{c}
		\includegraphics[scale=0.235]{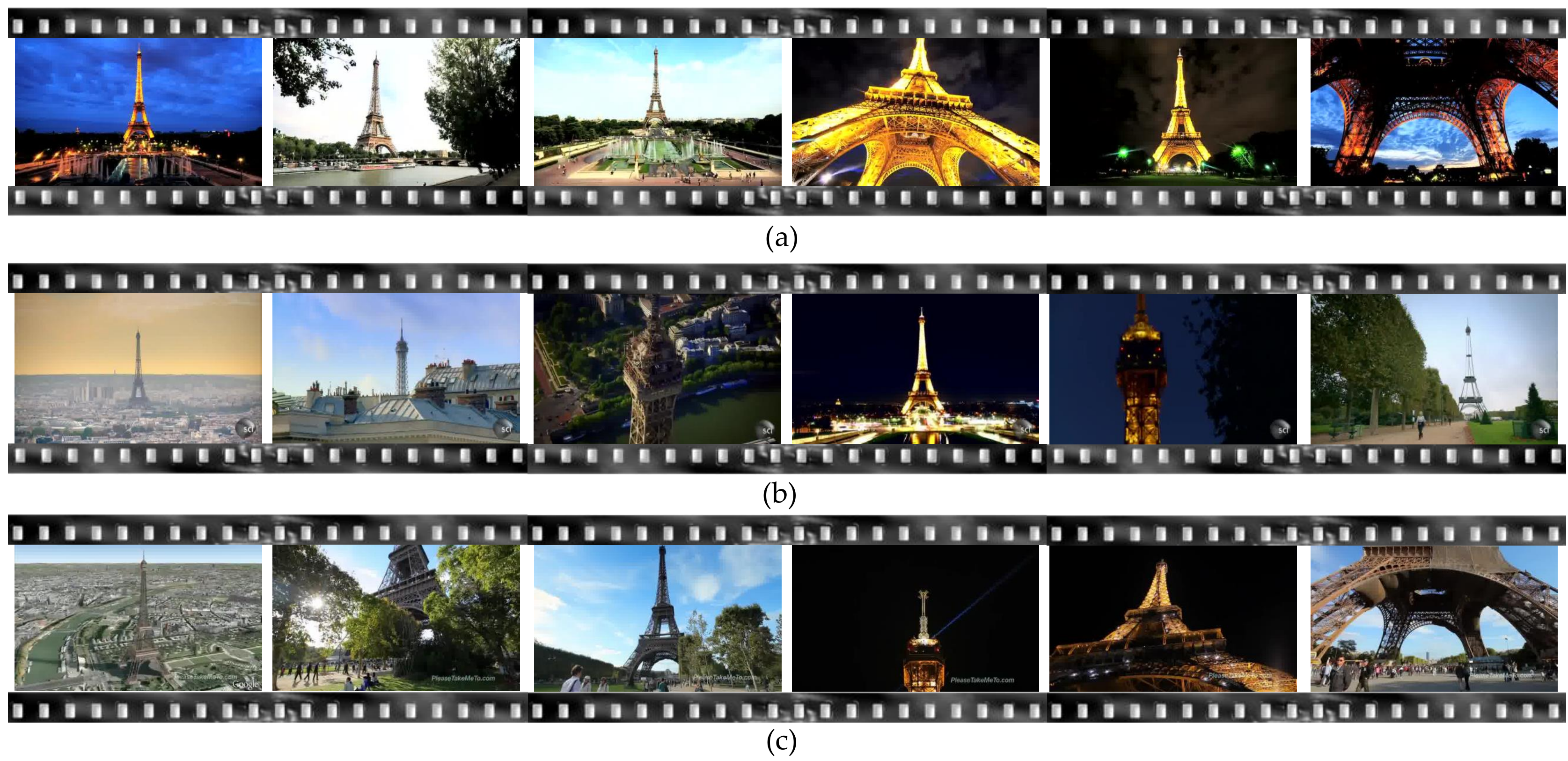}
	\end{tabular}
	\caption
	{Consider three videos of the topic \enquote{\textit{Eiffel Tower}}. Each row shows six uniformly sampled shots represented by the middle frame, from the corresponding video.
	It is clear that all these videos have mutual influence on each other since many visual concepts tend to appear repeatedly across them. We therefore hypothesize that such topically close videos can provide more knowledge and useful clues to extract summary from a given video.  
	We build on this intuition to propose a summarization algorithm that exploits topic-related visual context from video (b) \& (c) to automatically extract an informative summary from a given video (a).}
	\label{fig:MotivationFigure} 
\end{figure} 

Much progress has been made in developing a variety of ways to summarize videos, by exploring different design criteria (representativeness~\cite{Khosla2013,Ehsan2012,Eric2014,Scalable2012,song2015tvsum,chu2015video}, interestingness~\cite{Onlinecontent2012,Att2005,Webcam2007}, importance~\cite{LucVanGool2014,xu2015gaze}) in an unsupervised manner, or developing supervised algorithms~\cite{Graumann2012,gygli2015video,gong2014diverse,Category2014,Ranking2014}.
However, with the notable exception of~\cite{chu2015video}, one common assumption of existing methods is that videos are independent of each other, and hence the summarization tasks are conducted separately by neglecting relationships that possibly reside across the videos.

Let us consider the video in Fig. 1a. 
The video is represented by six uniformly sampled shots. 
Now consider the videos in Fig. 1b and 1c along with the video in Fig. 1a. 
Are these videos independent of each other or something common exists across them? 
The answer is clear: all of these videos belong to the same topic \enquote{\textit{Eiffel Tower}}. 
As a result, the summaries of these videos will have significant common information with each other. 
Thus, the context of additional topic-related videos can be beneficial by providing more knowledge and additional clues for extracting a more informative and compact summary from a specified video. 
We build on this intuition, presenting a new perspective to summarize a video by exploiting the neighborhood knowledge from a set of topic-related videos.

In this paper, we propose a \textit{Collaborative Video Summarization (CVS)} approach that exploits visual context from a set of topic-related videos to extract an informative summary of a given video. Our work builds upon the idea of collaborative techniques~\cite{balabanovic1997fab,liu2008output,xue2005scalable} from information retrieval (IR) and natural language processing (NLP), which typically use the attributes of other similar objects to predict the attribute of a given object. We achieve this by \textit{finding a sparse set of representative and diverse shots that simultaneously capture both important particularities arising in the given video, as well as, generalities identified from the set of topic-related videos.} 
Our underlying assumption is that a few topically close videos actually have mutual influence on each other since many important visual concepts tend to appear repeatedly across them. 

Our approach works as follows. 
First, we segment each video into multiple non-uniform shots using a temporal segmentation algorithm and represent each shot by a feature vector using a mean pooling scheme over the extracted C3D features (Section~\ref{sec:video representation}). 
Then, we develop a novel collaborative sparse representative selection strategy by exploiting visual context from topic-related videos (Section~\ref{sec:collaborative sparse}). 
Specifically, we formulate the task of finding summaries as an $\ell_{2,1}$-norm sparse optimization problem where the nonzero rows of a sparse coefficient matrix represent the relative importance of the corresponding shots. 
Finally, the approach outputs a video summary composed of the shots with the highest importance score (Section~\ref{sec:summary generation}).
Note that the summary will be of the one video of interest only, while exploiting visual context from additional topic-related videos\footnote{In this work, we assume that additional topic-related videos are available beforehand. However, in most practical cases, videos retrieved from search engines with topic name as a query may contain outliers and irrelevant videos due to inaccurate query text and polysemy. One feasible choice is to use either clustering~\cite{jing2006igroup} or additional meta data to refine the results.} 
 
The main \textbf{contributions} of our work are as follows:
\newline
$\bullet$ We propose a novel approach to extract an informative and diverese summary of a specified video by exploiting additional knowledge from topic-related videos. The additional topic-related videos provide visual context to identify what is important in a video. 
\newline
$\bullet$ We develop a collaborative sparse representative selection strategy by introducing a consensus regularizer that simultaneously captures both important particularities arising in the given video, as well as, generalities identified from the additional topic-related videos. 
\newline
$\bullet$ We present an efficient optimization algorithm based on half-quadratic function theory to solve the non-smooth objective, where the minimization problem is simplified to two independent
linear system problems. 
\newline
$\bullet$ We demonstrate the effectiveness of our approach in two video summarization tasks---topic-oriented video summarization and multi-video concept visualization.
With extensive experiments on both CoSum~\cite{chu2015video} and TVSum50~\cite{song2015tvsum} video datasets, we show the superiority of our approach over competing methods for both summarization tasks.
\vspace{-1mm}
\section{Related Work}
\vspace{-1mm}
Video summarization has been studied from multiple perspectives~\cite{money2008video,Truong2007}. 
While the approaches might be supervised or unsupervised, the goal of summarization is nevertheless to produce a compact visual summary that encapsulates the most informative parts of a video. 

Much work has been proposed to summarize a video using supervised learning. 
Representative methods use category-specific classifiers for importance scoring~\cite{Category2014,Ranking2014} or learn how to select informative and diverse video subsets from human-created summaries~\cite{gygli2015video,gong2014diverse,sharghi2016query,zhang2016video} or learn important facets, like faces, hands, objects~\cite{Graumann2012,lu2013story,Sigal2015}. 
Although these methods have shown impressive results, their performance largely depends on huge amount of labeled examples which are difficult to collect for unconstrained web videos. 
Our CVS approach, on the other hand, exploits visual context from topic-related videos without requiring any labeled examples, and thus can be easily applied to summarize large scale web videos with diverse content. 

Without supervision, summarization methods must rely on low-level visual indices to determine the relevance of parts of a video. Various strategies have been studied, including clustering~\cite{VISON2012,VSUMM2011,Top2014,panda2014scalable}, interest prediction~\cite{Att2005,LucVanGool2014}, and energy minimization~\cite{Peleg2007,Onlinecontent2012}. Leveraging crawled web images is also another recent trend for video summarization~\cite{Khosla2013,song2015tvsum,Joint2014}. 
However, all of these methods summarize videos independently by neglecting relationships that possibly reside across them. The use of neighboring topic-related videos to improve summarization still remains as a novel and largely under-addressed problem.

The most relevant work to ours is the video co-summarization approach (CoSum)~\cite{chu2015video}. 
It aims to find visually co-occurring shots across videos of the same topic based on the idea of commonality analysis~\cite{chu2012unsupervised}. 
CoSum also introduced a new benchmark dataset for topic-oriented video summarization. 
However, CoSum and our approach have significant differences. 
CoSum constructs weighted bipartite graphs for each pair of videos in order to find the maximal bicliques, which can be computationally inefficient given a large collection of topic-related videos.
Our approach, on the other hand, offers a more flexible way to find most representative and diverse video shots through a collaborative sparse optimization framework that can be efficiently solved to handle large number of web videos simultaneously.
In addition, CoSum employs a computationally-intensive shot-level feature representation, namely a combination of both observation and interaction features~\cite{hoai2011joint}, which involves extracting low-level features such as CENTRIST, Dense-SIFT and HSV color moments. 
By contrast, our approach utilizes generic deep learning features which are more computationally efficient and more accurate in characterizing both appearance and motion.  

Our focus on sparse coding as the building block of CVS is largely inspired by its appealing property in modeling sparsity and representativeness in data summarization. 
In contrast to prior works~\cite{Scalable2012,Ehsan2012,Eric2014,panda2016embedded,panda2016video}, we develop a novel collaborative sparse optimization that finds shots which are informative about the given video, as well as, the set of topic-related videos. 

In recent years, collaborative techniques have been successfully applied to several IR and NLP tasks: collaborative recommendation~\cite{balabanovic1997fab,sarwar2001item}, collaborative filtering~\cite{xue2005scalable}, collaborative ranking~\cite{balakrishnan2012collaborative} and text summarization~\cite{wan2007single,wan2010exploiting,wan2007collabsum}. 
The common idea underlying all of these works, including ours, is to make use of the interactions among multiple objects under the assumption that similar objects will have similar behaviors and characteristics.
 
\vspace{-2mm}    
\section{Collaborative Video Summarization}
\label{sec:cvs}
\vspace{-1.5mm}
A summary is a condensed synopsis that conveys the most \textit{important} details of the original video. 
Specifically, it is composed of several shots that represent most important portions of the input video within a short duration.   
Since, \textit{importance} is a subjective notion, we define a good summary as one that has the following properties.
\newline
$\bullet$ \textbf{Representative.} The original video should be reconstructed with high accuracy using the extracted summary. We extend this notion of representative as finding a summary that simultaneously minimizes reconstruction error of the given video, as well as the set of topic-related videos.
\newline
$\bullet$ \textbf{Sparsity.} Although the summary should be representative of the input video, the length should be as small as possible.
\newline
$\bullet$ \textbf{Diversity.} The summary should be collectively diverse capturing different aspects of the video---otherwise one can remove some of it without losing much information. 

The proposed approach, CVS, decomposes into three steps: i) video representation; ii) collaborative sparse representative selection; iii) summary generation. 
\vspace{-1.5mm}
\subsection{Video Representation}
\label{sec:video representation}
\vspace{-1mm}

\textbf{Temporal Segmentation.} 
Our approach starts with segmenting videos using an existing algorithm~\cite{chu2015video}. We segment each video into multiple non-uniform shots with an additional constraint to ensure that the number of frames within each shot lies in the range of [32,96]. The segmented shots serve as the basic units for feature extraction and subsequent processing to extract a video summary.

\textbf{Feature Representation.} 
Recent advancement in deep learning has revealed that features extracted from upper or intermediate layers of convolutional neural networks (CNNs) are generic features that have good transfer learning capabilities across different domains~\cite{simonyan2014two,zhou2014learning,karpathy2014large,russakovsky2015imagenet}.
In the case of videos, C3D features~\cite{tran2014c3d} have recently shown better performance compared to the features extracted using each frame separately~\cite{tran2014learning,yao2015describing}.
We therefore extract C3D features, by taking sets of 16 input frames, applying 3D convolutional filters, and extracting the responses at layer FC6 as suggested in~\cite{tran2014c3d}. 
This is followed by a temporal mean pooling scheme to maintain the local ordering structure within a shot. 
Then the pooling result serves as the final feature vector of a shot (4096 dimensional) to be used in the sparse optimization. We will discuss the performance benefits of employing C3D features later in our experiments. 
\vspace{-7mm}
\subsection{Collaborative Sparse Representative Selection}
\label{sec:collaborative sparse}
\vspace{-0.5mm}
We develop a sparse optimization framework that incorporates both information content of the given video and the topic-related videos to extract an informative summary of the specified video. Let \textit{v} be a video to be summarized and \textit{$\tilde{v}$} denote the set of remaining topic-related videos from the video collection. 
Let the feature matrix of the video \textit{v} and \textit{$\tilde{v}$} are given by $\textbf{X}\in\mathbb{R}^{d \times n}$and $\textbf{\~X}\in\mathbb{R}^{d \times \tilde{n}}$ respectively. $d$ is the dimensionality of the C3D features and $n$ represents the number of shots in the video $v$. $\tilde{n}$ represents the total number of shots in the remaining topic-related videos \textit{$\tilde{v}$}.

{\bf Formulation.} Sparse optimization approaches~\cite{Scalable2012,Ehsan2012} find the representative shots from a video itself by minimizing the linear reconstruction error as
\vspace{-2mm}
\begin{equation}
\small
\begin{gathered}
\label{eq:equation1}
\min_{\textbf{Z}\in \mathbb{R}^{n\times n}}\ \dfrac{1}{2}\lVert {\textbf{X} - \textbf{XZ} \rVert}^2_{F} + \lambda_s \lVert {\textbf{Z}\rVert}_{2,1} 
\end{gathered}
\vspace{-1mm}
\end{equation}
where $||\textbf{Z}||_{2,1} = \sum_{i=1}^{n} ||\textbf{Z}_{i}||_2$ and $||\textbf{Z}_{i}||_2$ is the $\ell_2$-norm of the $i$-th row of $\textbf{Z}$. $\lambda_s > 0 $ is a regularization parameter that controls the level of sparsity in the reconstruction. Once the problem (\ref{eq:equation1}) is solved, the representatives are selected as the points whose corresponding $||\textbf{Z}_i||_2 \neq 0$. 

Clearly, the above formulation summarizes a video neglecting mutual relationships that possibly reside across the videos. Considering the relationships across the topic-related videos, we aim to select a sparse set of representative shots that balances two main objectives: (i) they are informative about the given video, and (ii) they are informative about the complete set of topic-related videos. 
In other words, we aim to extract a summary that simultaneously minimizes the reconstruction error of the specified video, as well as the set of topic-related videos. 
Given the above stated goals, we formulate the following objective function,
\vspace{-2mm}
\begin{equation}
\small
	\begin{gathered}
		\label{eq:equation3}
		\min_{\textbf{Z,\ \~Z}}\ \dfrac{1}{2} 
		\big(\lVert {\textbf{X} - \textbf{XZ} \rVert}^2_{F} + \alpha \lVert {\textbf{\~X} - \textbf{X\~Z} \rVert}^2_{F}\big) \\
		+ \lambda_s \big(\lVert {\textbf{Z}\rVert}_{2,1} + \lVert {\textbf{\~Z}\rVert}_{2,1}\big)
	\end{gathered}
	\vspace{-1mm}
\end{equation} 
where parameter $\alpha > 0$ balances the penalty between errors in the reconstruction of video $v$ and errors in the reconstruction of the remaining videos in the collection \textit{$\tilde{v}$}\footnote{Note that we use a common $\alpha$ to weight the reconstruction term related to the topic-related videos in (\ref{eq:equation3}) for simplicity of exposition.  However, if we have some prior information on which video is more informative about the topic or close to the specified video, we can assign different $\alpha$s for different topic-related videos. We leave this problem about the different choice of $\alpha$ as an interesting future work.}. The objective function is intuitive: minimization of (\ref{eq:equation3}) favors selecting a sparse set of representative shots that simultaneously reconstructs the target video $\textbf{X}$ via $\textbf{Z}$, as well as the set of topic-related videos $\textbf{\~X}$ via $\textbf{\~Z}$, with high accuracy.

{\bf Diversity Regularization.}
The data reconstruction and sparse optimization formulations in (\ref{eq:equation3}) tend to select shots that can cover a specified video, as well as the set of topic-related videos. However, there is no explicit tendency to select diverse shots capturing different but also important information described in the set of videos. Prior works~\cite{Scalable2012,Ehsan2012} handle this issue by manually filtering redundant shots from the extracted summary which can be unreliable while summarizing large scale web videos. Recent works on sparse representative selection~\cite{imagecollection2012,wang2016scalable} also addresses this diversity problem by explicitly adding non-convex regularizers in the objective which makes it difficult to optimize.

Inspired by the recent work on convex formulation for active learning~\cite{6751135} and document compression~\cite{yao2015compressive}, we introduce two diversity regularization functions, $f_{d}{\textbf{(Z)}}$ and $f_{d}{\textbf{(\~Z)}}$ to select a sparse set of representative and diverse shots from the video. Our motivation is that rows in sparse coefficient matrices corresponding to two similar shots are not nonzero at the same time. This is logical since the representative shots should be non-redundant capturing diverse aspects of the input video.

{\bf Definition 1.}
Given the sparse coefficient matrices $\textbf{Z}$ and $\textbf{\~Z}$, the diversity regularization functions are defined as:
\vspace{-2.5mm}
\begin{equation}
\small
	\begin{gathered}
		\label{eq:divequation1}
		f_{d}{\textbf{(Z)}} = \sum_{i=1}^{n} \sum_{j=1}^{n} d_{ij}Z_{ij} = tr(\textbf{D}^{T}\textbf{Z}), \\ 
		f_{d}{\textbf{(\~Z)}} = \sum_{i=1}^{n} \sum_{j=1}^{\textit{$\tilde{n}$}} \textit{$\tilde{d}$}_{ij}\textit{$\tilde{Z}$}_{ij} = tr(\textbf{\~D}^{T}\textbf{\~Z})
	\end{gathered}
	\vspace{-2mm}
\end{equation}             
where $\textbf{D}$ is the weight matrix measuring the pair-wise similarity of shots in $\textbf{X}$, and $\textbf{\~D}$ measures the similarity between shots in $\textbf{X}$ and $\textbf{\~X}$. There are a lot of ways to construct $\textbf{D}$ and $\textbf{\~D}$. In this paper, we employ the inner product to measure the similarity, since it is simple to implement and it performs well in practice. Minimization of these functions tries to select diverse shots by penalizing the condition that rows of two similar shots are nonzero at the same time.

After adding the diversity regularization functions into problem (\ref{eq:equation3}), we have the objective function as follows:
\vspace{-2mm}
\begin{equation}
\small
\begin{gathered}
\label{eq:equation4}
\min_{\textbf{Z,\ \~Z}}\ \dfrac{1}{2} 
\big(\lVert {\textbf{X} - \textbf{XZ} \rVert}^2_{F} + \alpha \lVert {\textbf{\~X} - \textbf{X\~Z} \rVert}^2_{F}\big) \\
+ \lambda_s \big(\lVert {\textbf{Z}\rVert}_{2,1} + \lVert {\textbf{\~Z}\rVert}_{2,1}\big) + \lambda_d \big( tr(\textbf{D}^{T}\textbf{Z}) + tr(\textbf{\~D}^{T}\textbf{\~Z})\big)
\end{gathered}
\end{equation}
where $\lambda_d$ is a trade-off factor associated with the functions. 

{\textbf{Consensus Regularization.}} The objective function (\ref{eq:equation4}) favors selecting a sparse set of representative and diverse shots from a target video $\textbf{X}$ by exploiting visual context from additional topic-related videos $\textbf{\~X}$. Specifically, rows in $\textbf{Z}$ provide information on relative importance of each shot in describing the video $\textbf{X}$, while rows in $\textbf{\~Z}$ give information on relative importance of each shot in $\textbf{X}$ in describing $\textbf{\~X}$. Given the two sparse coefficient matrices, our next goal is to select a unified set of shots that simultaneously cover the important particularities arising in the target video, as well as the generalities arising in the video collection. 

To achieve the above goal, we propose to minimize the following objective function:
\vspace{-2mm}
\begin{equation}
\small
\begin{gathered}
\label{eq:equation5}
\min_{\textbf{Z,\ \~Z}}\ \dfrac{1}{2} 
\big(\lVert {\textbf{X} - \textbf{XZ} \rVert}^2_{F} + \alpha \lVert {\textbf{\~X} - \textbf{X\~Z} \rVert}^2_{F}\big) \\
+ \lambda_s \big(\lVert {\textbf{Z}\rVert}_{2,1} + \lVert {\textbf{\~Z}\rVert}_{2,1}\big) + \lambda_d \big( tr(\textbf{D}^{T}\textbf{Z}) + tr(\textbf{\~D}^{T}\textbf{\~Z})\big) \\ + \beta||\textbf{Z}_c||_{2,1} \ \ \ \  s.t. \ \  \textbf{Z}_c = [\textbf{Z}\vert\textbf{\~Z}], \ \textbf{Z}_c\in \mathbb{R}^{n\times (n+\tilde{n})}
\end{gathered}
\vspace{-1mm}
\end{equation}
where $\ell_{2,1}$-norm on the consensus matrix $\textbf{Z}_c$ enables $\textbf{Z}$ and $\textbf{\~Z}$ to have the similar sparse patterns and share the common components. The joint $\ell_{2,1}$-norm plays the role of consensus regularization as follows. In each round of the optimization algorithm
developed later in this paper, the updated sparse coefficient matrices in the former rounds can be used to regularize the current
optimization criterion. Thus, it can uncover the shared knowledge of $\textbf{Z}$ and $\textbf{\~Z}$ by suppressing irrelevant or noisy video shots, which results in an optimal $\textbf{Z}_c$ for selecting representative video shots.

{\bf Optimization.} 
Since problem (\ref{eq:equation5}) is non-smooth involving multiple $\ell_{2,1}$-norms, it is difficult to optimize directly. Half-quadratic optimization techniques~\cite{he20122,he2014half} have shown to be effective in solving these sparse optimizations in several computer vision applications~\cite{wang2013learning,peng2016robust,wang2015robust,lu2013correntropy,bergmann2015restoration}. 
Motivated by such methods, we devise an iterative algorithm to efficiently solve (\ref{eq:equation5}) by minimizing its augmented function alternatively.      
Specifically, if we define $\phi(x)=\sqrt{x^2+\epsilon}$ with $\epsilon$ being a constant, we can transform $\lVert {\textbf{Z}\rVert}_{2,1}$ to $\sum_{i=1}^{n}\sqrt{||\textbf{Z}_{i}||_2^2+\epsilon}$, according to the analysis of $\ell_{2,1}$-norm in~\cite{he20122,lu2013correntropy}. With this transformation, we can optimize (\ref{eq:equation5}) efficiently in an alternative way as follows. 

According to the half-quadratic theory~\cite{he20122,he2014half,geman1992constrained}, the augmented cost-function of (\ref{eq:equation5}) can be written as follows. 
\vspace{-2mm}
\begin{equation}
\small
	\begin{gathered}
		\label{eq:equation6}
		\min_{\textbf{Z,\ \~Z}}\ \dfrac{1}{2} 
		\big(\lVert {\textbf{X} - \textbf{XZ} \rVert}^2_{F} + \alpha \lVert {\textbf{\~X} - \textbf{X\~Z} \rVert}^2_{F}\big) \\
		+ \lambda_s \big(tr(\textbf{Z}^T\textbf{P}\textbf{Z})  + tr(\textbf{\~Z}^T\textbf{Q}\textbf{\~Z})\big) + \lambda_d \big( tr(\textbf{D}^{T}\textbf{Z}) + tr(\textbf{\~D}^{T}\textbf{\~Z})\big) \\ + \beta \big(tr(\textbf{Z}_c^T\textbf{R}\textbf{Z}_c)\big)
	\end{gathered}
	\vspace{-1mm}
\end{equation}
where $\textbf{P,\ Q,\ R}\in \mathbb{R}^{n\times n}$ are three diagonal matrices, and the corresponding $i$-th element is defined as
\begin{equation}
\small
\begin{gathered}
\label{eq:diagmat2}
\textbf{P}_{ii}=\dfrac{1}{2\sqrt{||\textbf{Z}_{i}||^2_2+\epsilon}}, \ \ \ \ 
\textbf{Q}_{ii}=\dfrac{1}{2\sqrt{||\textbf{\~Z}_{i}||^2_2+\epsilon}}, \\
\textbf{R}_{ii}=\dfrac{1}{2\sqrt{||{\textbf{Z}_c}_{i}||^2_2+\epsilon}}
\end{gathered}
\vspace{-1mm}
\end{equation}
where $\epsilon$ is a smoothing term, which is usually set to be a small constant value. Optimizing (\ref{eq:equation6}) over $\textbf{Z}$ and $\textbf{\~Z}$ is equivalent to optimizing the following two problems.
\vspace{-2mm}
\begin{equation}
\small
\begin{gathered}
\label{eq:equation7}
\min_{\textbf{Z}}\ \dfrac{1}{2} 
\lVert {\textbf{X} - \textbf{XZ} \rVert}^2_{F} 
+ \lambda_d tr(\textbf{D}^{T}\textbf{Z}) \\+ \lambda_s tr(\textbf{Z}^T\textbf{P}\textbf{Z}) + \beta tr(\textbf{Z}^T\textbf{R}\textbf{Z}) 
\end{gathered}
\vspace{-1mm}
\end{equation}
\begin{equation}
\small
\begin{gathered}
\label{eq:equation8}
\min_{\textbf{\~Z}}\ \dfrac{\alpha}{2}  \lVert {\textbf{\~X} - \textbf{X\~Z} \rVert}^2_{F} + \lambda_d tr(\textbf{\~D}^{T}\textbf{\~Z}) \\
+ \lambda_s tr(\textbf{\~Z}^T\textbf{Q}\textbf{\~Z}) + \beta tr(\textbf{\~Z}^T\textbf{R}\textbf{\~Z})
\end{gathered}
\end{equation}
Now with fixed $\textbf{P,\ Q,\ R}$, the optimal solution of (\ref{eq:equation7}) and (\ref{eq:equation8}) can be computed by solving the following linear systems:
\begin{equation}
\small
\begin{gathered}
\label{eq:equation9}
(\textbf{X}^T\textbf{X}+2\lambda_s \textbf{P}+2\beta\textbf{R})\textbf{Z}=(\textbf{X}^T\textbf{X}-\lambda_d\textbf{D}) \\
(\alpha\textbf{X}^T\textbf{X}+2\lambda_s \textbf{Q}+2\beta\textbf{R})\textbf{\~Z}=(\alpha\textbf{X}^T\textbf{\~X}-\lambda_d\textbf{\~D})
\end{gathered}
\vspace{-1mm} 
\end{equation}

\begin{algorithm} 
	\caption{Algorithm for Solving Problem (\ref{eq:equation5})}\label{algo:Algorithm}
	\begin{algorithmic}
		\State {\bf Input:} Video feature matrices $\textbf{X}$ and $\textbf{\~X}$; \\
		\ \ \ \ \ \ \ \ \ \ \ \  Parameters $\alpha,\lambda_s,\lambda_d,\beta$, set $t=0$;\\
		\ \ \ \ \ \ \ \ \ \ \ \ \ \ Construct $\textbf{D}$ and $\textbf{\^D}$ using inner product similarity; \\
		\ \ \ \ \ \ \ \ \ \ \ \  Initialize $\textbf{Z}$ and $\textbf{\~Z}$ randomly, set $\textbf{Z}_c$= [$\textbf{Z}$,\ $\textbf{\~Z}$] ;
		\State {\bf Output:} Optimal sparse coefficient matrix $\textbf{Zc}$.  
		\While{\textit{not converged}} \\
		 \ \ \  1. Compute $\textbf{P}^{t}$, $\textbf{Q}^{t}$ and $\textbf{R}^{t}$ using (\ref{eq:diagmat2}); \\
		\ \ \ 2. Compute $\textbf{Z}^{t+1}$ and $\textbf{\~Z}^{t+1}$ using (\ref{eq:equation9}); \\
		 \ \ \ 3.  Compute $\textbf{Z}_c^{t+1}$ as: $\textbf{Z}_c^{t+1}$= [$\textbf{Z}^{t+1}$ $\vert$ $\textbf{\~Z}^{t+1}$]; \\
		\ \ \ 4. $t=t+1$;
		\EndWhile
	\end{algorithmic} 
\end{algorithm}	
Algo.~\ref{algo:Algorithm} summarizes the alternative minimization procedure to optimize (\ref{eq:equation5}). In step 1, we compute the auxiliary matrices $\textbf{P}$, $\textbf{Q}$ and $\textbf{R}$ which play an important role in representative selection, according to the half-quadratic analysis for $\ell_{2,1}$-norm~\cite{he20122}. In step 2, we find the optimal sparse coefficient matrices $\textbf{Z}$ and $\textbf{\~Z}$ by solving two linear systems as defined in (\ref{eq:equation9}). Step 3 corresponds to the consensus matrix, which is expected to uncover the shared knowledge of $\textbf{Z}$ and $\textbf{\~Z}$ by enforcing same sparse pattern using a joint $\ell_{2,1}$-norm.       
 
\vspace{-1mm}      
\subsection{Summary Generation}
\label{sec:summary generation}
\vspace{-1mm}
Above, we described how we compute the optimal sparse coefficient matrix $\textbf{Z}_c$ by exploiting visual context from the topic-related videos. 
To generate a summary, we first sort the shots by decreasing importance according to the $\ell_{2}$ norms of the rows in $\textbf{Z}_c$ (resolving ties by favoring shorter video shots), and then construct the optimal summary from the top-ranked shots that fit in the length constraint.

\vspace{-2mm} 
\section{Experiments}
\label{sec:Experiments}
\vspace{-1mm} 
{\bf Datasets.} We evaluate the performance of our approach using two datasets: (i) the CoSum dataset~\cite{chu2015video} and (ii) the TVSum50 dataset~\cite{song2015tvsum}. To the best of our knowledge, these are the only two publicly available summarization datasets of multiple videos organized into groups with a topic keyword. Both of the datasets are extremely diverse: while CoSum dataset consists of 51 videos covering 10 topics from the SumMe benchmark~\cite{LucVanGool2014}, the TVSum50 dataset contains 50 videos organized into 10 topics from the TRECVid Multimedia Event Detection task~\cite{smeaton2006evaluation}.

\begin{table*} [t]
	\centering
	\scriptsize
	\caption{Experimental results on CoSum dataset. Numbers show top-5 AP scores averaged over all the videos of the same topic. We highlight the \textbf{best} and \underline{second best} baseline method. Overall, our approach outperforms all the baseline methods.}
	\label{tab:CoSum}
	\begin{tabulary}{1.1\linewidth}{|p{28mm}||P{8mm}|P{8mm}|P{8mm}||P{7.5mm}|P{7.5mm}|P{8.5mm}|P{8.2mm}|P{7.5mm}|P{8.5mm}|P{8mm}|}
		\hline
		& \multicolumn{3}{c||}{\textbf{\texttt{Humans}}} &\multicolumn{7}{c|}{\textbf{\texttt{Computational methods}}}\\
		\cline{2-11}
		\textbf{Video Topics} &\textbf{\texttt{Worst}} & \textbf{\texttt{Mean}} &\textbf{\texttt{Best}}& \textbf{\texttt{CK}} & \textbf{\texttt{CS}} & \textbf{\texttt{SMRS}} & \textbf{\texttt{LL}} & \textbf{\texttt{CoC}} & \textbf{\texttt{CoSum}} & \textbf{\texttt{CVS}} \\
		\hhline{|=|=|=|=|=|=|=|=|=|=|=|}
		Base Jumping	& 0.652	& 0.831	& 0.896 & 0.415 & 0.463	& 0.487 & 0.504 & 0.561 & \underline{0.631}	& \textbf{0.658}\\
		Bike Polo 	& 0.661	& 0.792	& 0.890 & 0.391 & 0.457	& 0.511 & 0.492 & \underline{0.625} & 0.592	& \textbf{0.675}\\
		Eiffel Tower 	& 0.697	& 0.758	& 0.881 & 0.398 & 0.445	& 0.532 & 0.556 & 0.575 & \underline{0.618}	& \textbf{0.722}\\
		Excavators River Xing 	& 0.705	& 0.814	& 0.912 & 0.432 & 0.395	& 0.516 & 0.525 & 0.563 & \underline{0.575}	& \textbf{0.693}\\
		Kids Playing in Leaves 	& 0.679	& 0.746	& 0.863 & 0.408 & 0.442	& 0.534 & 0.521 & 0.557 & \underline{0.594}	& \textbf{0.707}\\
		MLB 	& 0.698	& 0.861	& 0.914 & 0.417 & 0.458	& 0.518 & 0.543 & 0.563 & \underline{0.624}	& \textbf{0.679}\\
		NFL 	& 0.660	& 0.775	& 0.865 & 0.389 & 0.425	& 0.513 & 0.558 & 0.587 & \underline{0.603}	& \textbf{0.674}\\
		Notre Dame Cathedral 	& 0.683	& 0.825	& 0.904 & 0.399 & 0.397	& 0.475 & 0.496 & \underline{0.617} & 0.595	& \textbf{0.702}\\
		Statue of Liberty 	& 0.687	& 0.874	& 0.921 & 0.420 & 0.464	& 0.538 & 0.525 & 0.551 & \underline{0.602}	& \textbf{0.715}\\
		Surfing 	& 0.676	& 0.837	& 0.879 & 0.401 & 0.415	& 0.501 & 0.533 & 0.562 & \underline{0.594}	& \textbf{0.647}\\
		\hline 		
	\end{tabulary} 
	
	\begin{tabulary}{1.1\linewidth}{|p{28mm}||P{8mm}|P{8mm}|P{8mm}||P{7.5mm}|P{7.5mm}|P{8.5mm}|P{8.2mm}|P{7.5mm}|P{8.5mm}|P{8mm}|}
		\hline
		\textbf{mean} & \bf{0.679}	& \bf{0.812}	& \bf{0.893} & \bf{0.407} & \bf{0.436}	& \bf{0.511} & \bf{0.525} & \bf{0.576} & \bf{0.602}	& \bf{0.687}\\
		\hhline{|-|-|-|-||-|-|-|-|-|-|-|}
		relative to average human 	& 83\%	& 100\%	& 110\% & 51\% & 54\%	& 62\% & 64\% & 70\% & 74\%	& 85\%\\
		\hline 		
	\end{tabulary} 
	\vspace{-1mm}
\end{table*} 

\begin{table*} [t]
	\centering
	\scriptsize
	\caption{Experimental results on TVSum50 dataset.
		}
	\label{tab:TVSUM50}
	\begin{tabulary}{1.1\linewidth}{|p{28mm}||P{8mm}|P{8mm}|P{8mm}||P{7.5mm}|P{7.5mm}|P{8.5mm}|P{8.2mm}|P{7.5mm}|P{8.5mm}|P{8mm}|}
		\hline
		& \multicolumn{3}{c||}{\textbf{\texttt{Humans}}} &\multicolumn{7}{c|}{\textbf{\texttt{Computational methods}}}\\
		\cline{2-11}
		\textbf{Video Topics} &\textbf{\texttt{Worst}} & \textbf{\texttt{Mean}} &\textbf{\texttt{Best}}& \textbf{\texttt{CK}} & \textbf{\texttt{CS}} & \textbf{\texttt{SMRS}} & \textbf{\texttt{LL}} & \textbf{\texttt{CoC}} & \textbf{\texttt{CoSum}} & \textbf{\texttt{CVS}} \\
		\hhline{|=|=|=|=|=|=|=|=|=|=|=|}
		Changing Vehicle Tire	& 0.285	& 0.461	& 0.589 & 0.225 & 0.235	& 0.287 & 0.272 & \textbf{0.336} & 0.295	& \underline{0.328}\\
		Getting Vehicle Unstuck 	& 0.392	& 0.505	& 0.634 & 0.248 & 0.241	& 0.305 & 0.324 & \underline{0.369} & 0.357	& \textbf{0.413}\\
		Grooming an Animal 	& 0.402	& 0.521	& 0.627 & 0.206 & 0.249	& 0.329 & 0.331 & \underline{0.342} & 0.325	& \textbf{0.379}\\
		Making Sandwich 	& 0.365	& 0.507	& 0.618 & 0.228 & 0.302	& 0.366 & 0.362 & 0.375 & \textbf{0.412}	& \underline{0.398}\\
		ParKour 	& 0.372	& 0.503	& 0.622 & 0.196 & 0.223	& 0.311 & 0.289 & \underline{0.324} & 0.318	& \textbf{0.354}\\
		PaRade 	& 0.359	& 0.534	& 0.635 & 0.179 & 0.216	& 0.247 & 0.276 & 0.301 & \underline{0.334}	& \textbf{0.381}\\
		Flash Mob Gathering 	& 0.337	& 0.484	& 0.606 & 0.218 & 0.252	& 0.294 & 0.302 & 0.318 & \underline{0.365}	& \textbf{0.365}\\
		Bee Keeping	& 0.298	& 0.515	& 0.591 & 0.203 & 0.247	& 0.278 & 0.297 & 0.295 & \underline{0.313}	& \textbf{0.326}\\
		Attempting Bike Tricks 	& 0.365	& 0.498	& 0.602 & 0.226 & 0.295	& 0.318 & 0.314 & 0.327 & \underline{0.365}	& \textbf{0.402}\\
		Dog Show 	& 0.386	& 0.529	& 0.614 & 0.187 & 0.232	& 0.284 & 0.295 & 0.309 & \underline{0.357}	& \textbf{0.378}\\
		\hline 		
	\end{tabulary} 
	
	\begin{tabulary}{1.1\linewidth}{|p{28mm}||P{8mm}|P{8mm}|P{8mm}||P{7.5mm}|P{7.5mm}|P{8.5mm}|P{8.2mm}|P{7.5mm}|P{8.5mm}|P{8mm}|}
		\hline
		\textbf{mean} & \bf{0.356}	& \bf{0.505}	& \bf{0.613} & \bf{0.211} & \bf{0.249}	& \bf{0.301} & \bf{0.306} & \bf{0.329} & \bf{0.345}	& \bf{0.372}\\
		\hhline{|-|-|-|-||-|-|-|-|-|-|-|}
		relative to average human 	& 71\%	& 100\%	& 121\% & 42\% & 49\%	& 60\% & 61\% & 65\% & 68\%	& 74\%\\
		\hline 		
	\end{tabulary} \vspace{-2mm}	
\end{table*} 
    
{\bf Implementation details.} 
Our results can be reproduced through the following parameters. The regularization parameters $\lambda_s$ and $\beta$ are taken as $\lambda_{0}/\gamma$ where $\gamma > 1$ and $\lambda_{0}$ is analytically computed from the data~\cite{Ehsan2012}. The other parameters $\alpha$ and $\lambda_d$ are empirically set to 0.5 and 0.01 respectively and kept fixed for all results. 
\begin{table*}
	\centering
	\scriptsize
	\caption{Performance comparison between 2D CNN(VGG) and 3D CNN(C3D) features. Numbers show top-5 AP scores averaged over all the videos of the same topic. * abbreviates topic name for display convenience. See Tab.~\ref{tab:CoSum} for full names. 
	}
	\label{tab:FeatureComp}
	\begin{tabulary}{1.1\linewidth}{|p{21.9mm}||P{7mm}|P{7mm}|P{7.5mm}|P{14mm}|P{7mm}|P{6mm}|P{6mm}|P{8mm}|P{9mm}|P{8mm}|P{7mm}|}
		\hline
		\textbf{Methods} &\textbf{Base*} & \textbf{Bike*} &\textbf{Eiffel*}& \textbf{Excavators*} & \textbf{Kids*} & \textbf{MLB} & \textbf{NFL} & \textbf{Notre*} & \textbf{Statue*} & \textbf{Surfing} & \textbf{mean} \\
		\hhline{|=|=|=|=|=|=|=|=|=|=|=|=|}
		\texttt{CVS(Features\cite{chu2015video})} 	& 0.580	& 0.632	& 0.677 & 0.614 & 0.598	& 0.607 & 0.575 & 0.612 & 0.655	& 0.623 & 0.618\\
		\texttt{CVS(VGG)}	& 0.591	& 0.626	& 0.724 & 0.638 & 0.617	& 0.642 & 0.615 & 0.604 & 0.721	& 0.649& \underline{0.643}\\
		\texttt{CVS(C3D)} 	& 0.658	& 0.675	& 0.722 & 0.693 & 0.707	& 0.679 & 0.674 & 0.702 & 0.715	& 0.647 & \textbf{0.687}\\
		\hline 		
	\end{tabulary} 
	\vspace{-2mm}
\end{table*}
\begin{table*}
	\centering
	\scriptsize
	\caption{Ablation analysis of the proposed approach with different constraints on (\ref{eq:equation5}).} 
	\label{tab:Ablation}
	\begin{tabulary}{1.1\linewidth}{|p{22mm}||P{6.9mm}|P{6.9mm}|P{7.5mm}|P{14mm}|P{7mm}|P{6mm}|P{6mm}|P{8mm}|P{9mm}|P{8mm}|P{6.8mm}|}
		\hline
		\textbf{Methods} &\textbf{Base*} & \textbf{Bike*} &\textbf{Eiffel*}& \textbf{Excavators*} & \textbf{Kids*} & \textbf{MLB} & \textbf{NFL} & \textbf{Notre*} & \textbf{Statue*} & \textbf{Surfing} & \textbf{mean} \\
		\hhline{|=|=|=|=|=|=|=|=|=|=|=|=|}
		\texttt{CVS-Neighborhood} 	& 0.552	& 0.543	& 0.551 & 0.583 & 0.510	& 0.529 & 0.534 & 0.532 & 0.516	& 0.527 & 0.538\\
		\texttt{CVS-Diversity} 	& 0.643	& 0.650	& 0.678 & 0.672 & 0.645	& 0.653 & 0.619 & 0.666 & 0.688	& 0.609& \underline{0.654}\\
		\texttt{CVS}	& 0.658	& 0.675	& 0.722 & 0.693 & 0.707	& 0.679 & 0.674 & 0.702 & 0.715	& 0.647 & \textbf{0.687}\\
		\hline 		
	\end{tabulary} 
	\vspace{-1mm}
\end{table*}

{\bf Compared methods.} We compare our approach to the following baselines. For all of the methods, we use what is recommended in the published work. 

\textbf{Clustering (\texttt{CK} and \texttt{CS}):} 
We first clustered the shots using \textit{k}-means (\texttt{CK}) and spectral clustering (\texttt{CS}), with \textit{k} set to 20~\cite{chu2015video}.
We then generate a summary by selecting shots that are closest to the centroid of top largest clusters. 

\textbf{Sparse Coding (\texttt{SMRS} and \texttt{LL}):} We tested two approaches: Sparse Modeling Representative Selection (\texttt{SMRS})~\cite{Ehsan2012} and LiveLight (\texttt{LL})~\cite{Eric2014}.
\texttt{SMRS} finds the representative shots using the entire video as the dictionary and selecting key shots based on the zero patterns of the coding vector. Note that~\cite{Scalable2012} also uses the same objective function as in~\cite{Ehsan2012} for summarizing consumer videos. 
The only difference lies in the algorithm used to solve the objective function
(Proximal vs ADMM). Hence, we compared only with~\cite{Ehsan2012}. 
LL generates a summary over time by measuring the redundancy using a dictionary of shots updated online. We implemented it using SPAMS library~\cite{mairal2010online} with dictionary of size 200 and the threshold $\epsilon_{0}=0.15$, as in~\cite{Eric2014}.

\textbf{Co-occurrence Statistics (\texttt{CoC} and \texttt{CoSum}):} We compared with two baselines that leverage visual co-occurrence across the topic-related videos to generate a summary. 
Co-clustering (\texttt{CoC})~\cite{dhillon2001co} generates a summary by partitioning the graph into co-clusters such that each cluster contains a subset of shot-pairs with high visual similarity. On the other hand, \texttt{CoSum} finds maximal bicliques from the complete bipartite graph using a block coordinate descent algorithm. We generate a summary by selecting top-ranked shots based on the visual co-occurrence score and set the threshold to select maximal bicliques to 0.3, following~\cite{chu2015video}.

All methods (including the proposed one) use the same C3D feature as described in Sec.~\ref{sec:video representation}. Such an experimental setting can give a fair comparison for various methods. 

\vspace{-1mm}    
\subsection{Topic-oriented Video Summarization}
\label{sec:Topic}
\vspace{-1mm}
\textbf{Goal:}  
\textit{Given a set of web videos sharing a common topic (e.g., Eiffel Tower), the goal is to provide the users with summaries of each video that are relevant to the topic.}

{\bf Solution.} The objective function (\ref{eq:equation5}) extracts summary of a specified video by exploiting the visual context of topic-related videos. Given a set of videos, our approach can find summaries of each video by exploiting the additional knowledge from the remaining videos. Moreover, one can easily parallelize the computation for more computational efficiency given our alternating minimization in Algo.~\ref{algo:Algorithm}. This provides scalability to our approach in processing large number of web videos simultaneously. 

{\bf Evaluation.} Motivated by~\cite{chu2015video,Khosla2013}, we assess the quality of an automatically generated summary by comparing it to human judgment. In particular, given a proposed summary and a set of human selected summaries, we compute the pairwise average precision (AP) and then report the mean value motivated by the fact that there exists not a single ground truth summary, but multiple summaries are possible. 
Average precision is a function of both precision and change in recall, where precision indicates how well all the representative shots match with the reference summaries and recall indicates how many and how accurately are the representative shots returned in the retrieval result.

For CoSum dataset, we follow~\cite{chu2015video} and compare each video summary with five human created summaries\footnote{The original CoSum dataset contains three human created summaries. We have added two more ground truth summaries which are collected using a similar experiment, as in~\cite{chu2015video}.
}, whereas for TVSum50 dataset, we compare each summary with twenty ground truth summaries that are created via crowdsourcing. Since the ground truth annotations in TVSum50 dataset contain frame-wise importance scores, we first compute the shot-level importance scores by taking average of the frame importance scores within each shot and then select top 50\% shots for each video, as in~\cite{chu2015video}. 

Apart from comparing with the baseline methods, we also compute the average precision between human created summaries. We show the worst, average and best scores of the human selections. The worst human score is computed using the summary which is the least similar to the rest of the summaries whereas the best score represent the most similar summary that contain most shots that were selected by many humans. This provides a pseudo-upper bound for this task, and thus we also report normalized AP scores by rescaling the mean AP of human selections to 100\%. 

\textbf{Comparison with baseline methods.} 
Tab.~\ref{tab:CoSum} shows the AP on top 5 shots included in the summaries for CoSum dataset. We can see that our method significantly outperforms all baseline methods to achieve an average performance of 85\%, while the closest published competitor, \texttt{CoSum}, reaches 74\%.
Moreover, if we compare to the human performance, we can see that our method even outperforms the \texttt{worst human} score of each topic in most cases. 
This indicates that our method produces summaries comparable to human created summaries.
Similarly, for the top-15 results, our approach achieved the highest average score of 83\% compared to 69\% by the \texttt{CoSum} baseline. 

\begin{figure*}[h!]
	\centering
	\begin{tabular}{c}
		\includegraphics[scale=0.285]{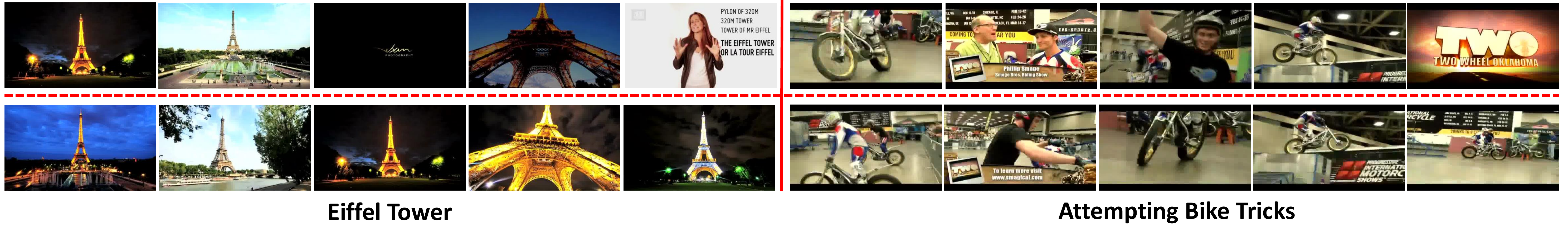}
	\end{tabular}
	\vspace{-2.5mm}
	\caption
	{  
		Role of topic-related visual context in summarizing a video. Top row: \texttt{CVS} w/o topic-related visual context, and Bottom row: \texttt{CVS} w/ topic-related visual context. 
		As can be seen, \texttt{CVS} w/o visual context often selects some shots that are irrelevant and not truly related to the topic.
		\texttt{CVS} w/ visual context, on the other hand, automatically selects the maximally informative shots by exploiting the information from additional neighborhood videos. Best viewed in color. 
	}
	\label{fig:SideFigure} 
	\vspace{-3mm}
\end{figure*} 
Our approach performed particularly well on videos that have their visual concepts described well by the topic-related videos, e.g., a video of the topic \textit{Eiffel Tower} contains shots that shows the night view of the tower and the remaining videos in the collection also depicts this well (Fig.~\ref{fig:MotivationFigure}).
While our method overall produces better summaries, it has a low performance for certain videos, e.g., videos of the topic \textit{Surfing}. These videos contain fast motion and subtle semantics that define representative shots of the video, such as surfing on the wave or sea swimming. We believe these are difficult to capture without an additional semantic analysis~\cite{mei2013near}; we leave this as future work.   

Tab.~\ref{tab:TVSUM50} shows top-5 AP results for the TVSum50 datset. Summarization in this dataset is more challenging because of the unconstrained topic keywords. Our approach still outperforms all the alternative methods significantly to achieve an average performance of 74\%. Similarly for top-15 results, our approach achieved highest score of 75\% compared to 66\% by the CoSum baseline.

\textbf{Test of Statistical Significance.} To show statistical significance, we have done t-test of our results and observe that the proposed approach, \texttt{CVS}, statistically significantly outperforms all six compared methods ($p<.01$), except for \texttt{worst human}. To further interpret the not-statistically significant result with respect to \texttt{worst human}, we perform a statistical power analysis ($\alpha = 0.01$) and see that the power computed for top-5 mAP results on CoSum dataset is 0.279, while on combining with top-15 results, it reaches to 0.877. Similarly, the power reaches 1 for a test that combines both top-5 and top-15 results of both of the datasets. Since, power of a high quality test should usually be $>0.80$, we can conclude that our approach statistically outperforms the \texttt{worst human} for a large sample size.     

\textbf{Effectiveness of C3D features.} We investigate the importance and reliability of C3D features by comparing with 2D shot-level deep features, and found that the later produces inferior results, with a top-5 mAP score of 0.643 on the CoSum dataset (Tab.~\ref{tab:FeatureComp}). We utilize Pycaffe~\cite{jia2014caffe} with the VGG net pretrained model~\cite{simonyan2014very} to extract a 4096-dim feature vector of a frame and then use temporal mean pooling to compute a single shot-level feature vector, similar to C3D features described in Sec.~\ref{sec:video representation}. We also compare with the shallow feature representation presented in~\cite{chu2015video} and observe that C3D features performs significantly better over shallow features in summarizing videos (0.618 vs 0.687). 
We believe this is because C3D features exploit the temporal aspects of activities typically shown in videos.  

\textbf{Performance of the individual components.} To better understand the contribution of various components in (\ref{eq:equation5}), we analyzed the performance of the proposed approach, by ablating each constraint while setting corresponding regularizer to zero (Tab.~\ref{tab:Ablation}). 
With all the components working, the mAP for the CoSum dataset is 0.687. By turning off the neighborhood information from topic-related videos, the mAP decreases to 0.538 (\texttt{CVS-Neighborhood}). This corroborates the fact that additional knowledge of topic-related videos help in extracting better summaries, closer to the human selection (see Fig.~\ref{fig:SideFigure} for qualtitative examples). Similarly, by turning off the diversity constraint, the mAP becomes 0.654 (\texttt{CVS-Diversity}). We can see that additional knowledge of topic-related videos contributes more than the diversity constraint in summarizing web videos.   
 
\vspace{-1mm}
\subsection{Multi-video Concept Visualization}
\label{sec:Concept} 
\textbf{Goal:} \textit{Given a set of topic-related videos, can we generate a single summary that describes the collection altogether?} Specifically, our goal is to generate a single video summary that better estimates human's visual concepts.

{\bf Solution.} A simple option would be to combine the individual summaries generated from Section.~\ref{sec:Topic} and select top ranked shots, regardless of the video, as in the existing existing method~\cite{chu2015video}. However, such choice will produce a lot of redundant events which eventually reduces the quality of the final summary. We believe this is because, although the individual summaries are informative and diverse, there exists redundancy across the extracted summaries that are relevant to the topic. Our approach can handle this by combining the summaries into a single video, say $\textbf{X}$ and then extracting a single diverse summary using the final objective function (\ref{eq:equation5}) with setting ($\alpha,\beta,\textbf{\~D}$) equal to zero. 
\begin{figure}[h!]
	\centering
	\begin{tabular}{c}
		\includegraphics[scale=0.40]{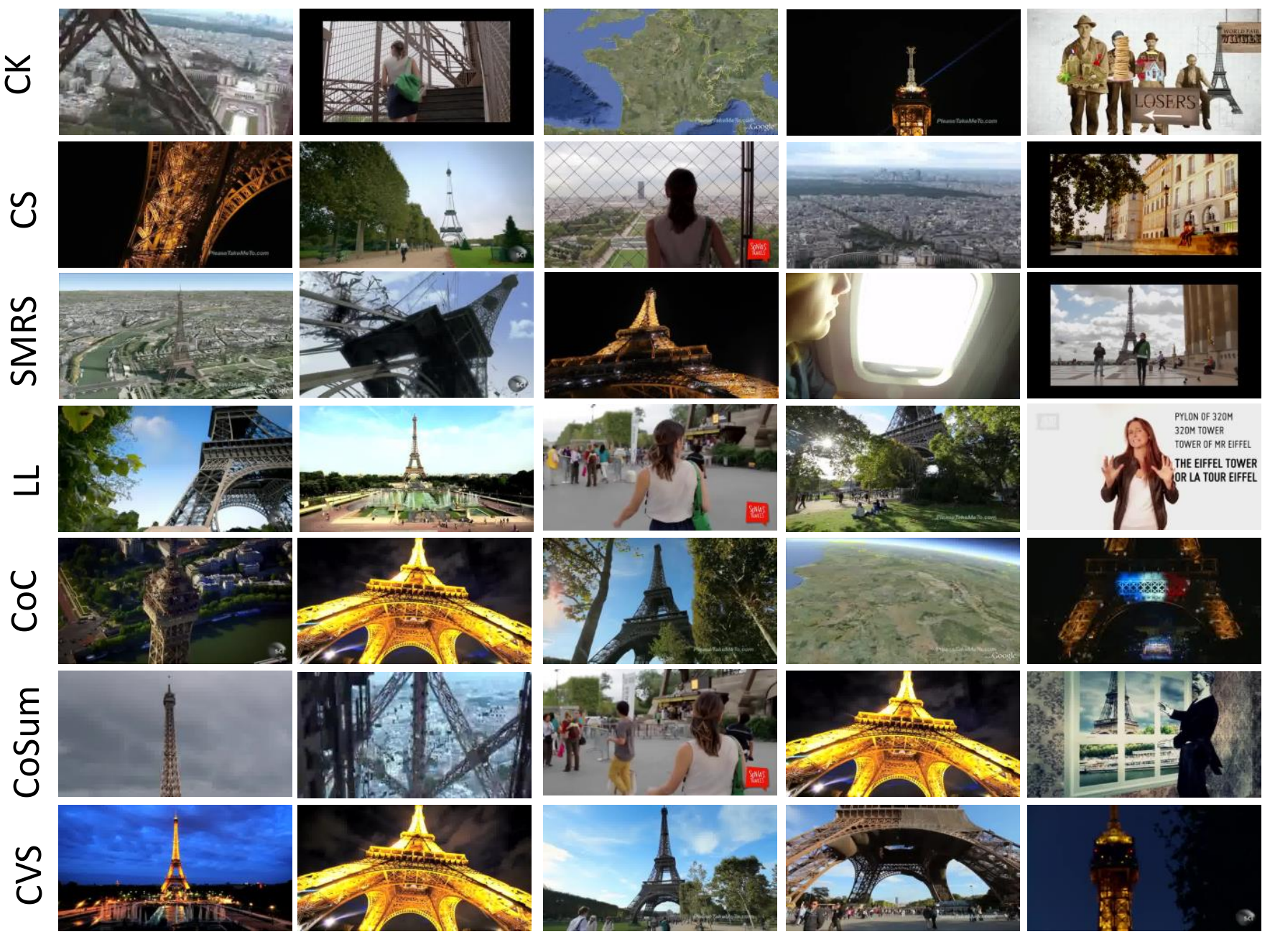}
	\end{tabular}
	\caption
	{  
		Illustrations of summaries constructed by different methods for the topic \textit{Eiffel Tower}. We show the top-5 results represented by the central frame of each shot. 
		Best viewed in color.       
	}
	\label{fig:SummaryFigure} 
\end{figure}
\begin{table}
	\centering
	\scriptsize
	\caption{\textbf{User Study---} Average expert ratings in concept visualization experiments. Our approach significantly outperforms other baseline methods in both of the datasets.
	}
	\label{tab:UserComp}
	\begin{tabulary}{1.1\linewidth}{|p{10mm}||P{4mm}|P{4mm}|P{6mm}|P{5mm}|P{5mm}|P{7mm}|P{6mm}|}
		\hline
		\textbf{Datasets} &\textbf{\texttt{CK}} & \textbf{\texttt{CS}} &\textbf{\texttt{SMRS}} & \textbf{\texttt{LL}} & \textbf{\texttt{CoC}} & \textbf{\texttt{CoSum}} & \textbf{\texttt{CVS}} \\
		\hhline{|=|=|=|=|=|=|=|=|}
		CoSum	& 3.70	& 4.03	& 5.60 & 5.63 & 6.64	& \underline{7.53} & \textbf{8.20} \\
		TVSum50 	& 2.46	& 3.06	& 4.02 & 4.20 & 4.8	& \underline{5.70} & \textbf{6.36}\\
		\hline 		
	\end{tabulary} 
\end{table} 

{\bf Evaluation.} To evaluate multi-video concept visualization, we need a single ground truth summary of all the topic-related videos that describes the collection altogether. However, since there exists no such ground truth summaries for both of the datasets, we performed human evaluations using 10 experts.
Given a video, the study experts were first shown the topic key word (\textit{e.g., Eiffel Tower}) and then shown the summaries constructed using different methods. They were asked to rate the overall quality of each summary by assigning a rating from 1 (worst) to 10 (best).   
 
{\bf Results.} 
Tab.~\ref{tab:UserComp} shows average expert ratings for both CoSum and TVSum50 datasets.
Similar to the results of topic-oriented summarization, our approach significantly outperforms all the baseline methods which indicates that our method generates a more informative summary that describes the video collection altogether. 
Furthermore, we note that the relative rank of the different approaches are largely preserved as compared to the topic-oriented summarization results.      
We show a visual comparison between the summaries produced by different methods in Fig.~\ref{fig:SummaryFigure}. As can be seen, our approach, \texttt{CVS}, generates a summary that better estimates human's visual concepts related to the topic. 

\vspace{-1mm}
\section{Conclusions}
\vspace{-1mm}
In this work, we present a novel video summarization framework that exploits visual context from a set of topic-related videos to extract an informative summary of a given video.
Motivated by the observation that important visual concepts tend to appear repeatedly across videos of the same topic, we develop a collaborative sparse optimization that finds a sparse set of representative and diverse shots by simultaneously capturing both important particularities arising in the given video, as well as, generalities arising across the video collection. 
We demonstrate the effectiveness of our approach on two standard datasets, significantly outperforming several baseline methods. 

\section*{\centering Appendix}
\label{sec:Convergence Analysis}
Since, we have solved (\ref{eq:equation5}) using an alternating minimization, we would like to show its convergence behavior. Specifically, the iterative approach in Algo.~\ref{algo:Algorithm} will monotonically decrease the objective value of (\ref{eq:equation5}) in each iteration.

\vspace{1mm}
As seen from (\ref{eq:equation6}), when we fix $\{\textbf{P},\textbf{Q},\textbf{R}\}$ as $\{\textbf{P}^t,\textbf{Q}^t,\textbf{R}^t\}$ in $t$-th iteration and compute $\textbf{Z}^{t+1}$, $\textbf{\~Z}^{t+1}$, $\textbf{Z}_c^{t+1}$, the following inequality holds,
\vspace{-1mm}
\begin{equation*}
\small
\begin{gathered}
\label{eq:proof0}
\dfrac{1}{2} 
\big(\lVert {\textbf{X} - \textbf{X}\textbf{Z}^{t+1} \rVert}^2_{F} + \alpha \lVert {\textbf{\~X} - \textbf{X}\textbf{\~Z}^{t+1} \rVert}^2_{F}\big) + \lambda_d tr(\textbf{D}^{T}\textbf{Z}^{t+1}) \\
+ \lambda_d tr(\textbf{\~D}^{T}\textbf{\~Z}^{t+1}) 
+ \lambda_s tr((\textbf{Z}^{t+1})^T\textbf{P}^{t}\textbf{Z}^{t+1}) \\ + \lambda_s tr((\textbf{\~Z}^{t+1})^T\textbf{Q}^{t}\textbf{\~Z}^{t+1})\big) + \beta \big(tr((\textbf{Z}_c^{t+1})^T\textbf{R}^{t}\textbf{Z}_c^{t+1})\big) 
\end{gathered}
\vspace{-1mm}
\end{equation*}
\vspace{-1mm}
\begin{equation}
\small
\begin{gathered}
\label{eq:proof1}
\leq
\dfrac{1}{2} 
\big(\lVert {\textbf{X} - \textbf{X}\textbf{Z}^{t} \rVert}^2_{F} + \alpha \lVert {\textbf{\~X} - \textbf{X}\textbf{\~Z}^{t} \rVert}^2_{F}\big) + \lambda_d tr(\textbf{D}^{T}\textbf{Z}^{t}) \\
+ \lambda_d tr(\textbf{\~D}^{T}\textbf{\~Z}^{t}) 
+ \lambda_s tr((\textbf{Z}^{t})^T\textbf{P}^{t}\textbf{Z}^{t}) \\ + \lambda_s tr((\textbf{\~Z}^{t})^T\textbf{Q}^{t}\textbf{\~Z}^{t})\big) + \beta \big(tr((\textbf{Z}_c^{t})^T\textbf{R}^{t}\textbf{Z}_c^{t})\big)
\end{gathered}
\end{equation}
Adding $\tiny \sum_{i=1}^{n}\frac{\epsilon}{2\sqrt{||\textbf{Z}_{i}^t||^2_2+\epsilon}}$ to both sides of (\ref{eq:proof1}), we have   
\begin{equation}
\small
\begin{gathered}
\label{eq:proof2}
\dfrac{1}{2} 
\big(\lVert {\textbf{X} - \textbf{X}\textbf{Z}^{t+1} \rVert}^2_{F} + \alpha \lVert {\textbf{\~X} - \textbf{X}\textbf{\~Z}^{t+1} \rVert}^2_{F}\big) + \lambda_d tr(\textbf{D}^{T}\textbf{Z}^{t+1}) \\
+ \lambda_d tr(\textbf{\~D}^{T}\textbf{\~Z}^{t+1}) 
+ \lambda_s \sum_{i=1}^{n}\dfrac{||\textbf{Z}_{i}^{t+1}||^2_2+\epsilon}{2\sqrt{||\textbf{Z}_{i}^t||^2_2+\epsilon}}  \\ + \lambda_s \sum_{i=1}^{n}\dfrac{||\textbf{\~Z}_{i}^{t+1}||^2_2+\epsilon}{2\sqrt{||\textbf{\~Z}_{i}^t||^2_2+\epsilon}} + \beta \sum_{i=1}^{n}\dfrac{||{\textbf{Z}_c}_{i}^{t+1}||^2_2+\epsilon}{2\sqrt{||{\textbf{Z}_c}_{i}^t||^2_2+\epsilon}} \\ \leq
\dfrac{1}{2} 
\big(\lVert {\textbf{X} - \textbf{X}\textbf{Z}^{t} \rVert}^2_{F} + \alpha \lVert {\textbf{\~X} - \textbf{X}\textbf{\~Z}^{t} \rVert}^2_{F}\big) + \lambda_d tr(\textbf{D}^{T}\textbf{Z}^{t}) \\
+ \lambda_d tr(\textbf{\~D}^{T}\textbf{\~Z}^{t}) 
+\lambda_s \sum_{i=1}^{n}\dfrac{||\textbf{Z}_{i}^{t}||^2_2+\epsilon}{2\sqrt{||\textbf{Z}_{i}^t||^2_2+\epsilon}} \\ + \lambda_s \sum_{i=1}^{n}\dfrac{||\textbf{\~Z}_{i}^{t}||^2_2+\epsilon}{2\sqrt{||\textbf{\~Z}_{i}^t||^2_2+\epsilon}} + \beta \sum_{i=1}^{n}\dfrac{||{\textbf{Z}_c}_{i}^{t}||^2_2+\epsilon}{2\sqrt{||{\textbf{Z}_c}_{i}^t||^2_2+\epsilon}}
\end{gathered}
\end{equation}
According to the \textit{Lemma} in~\cite{nie2010efficient}: 
\vspace{-1mm}
\begin{equation}
\small
\begin{gathered}
\label{eq:lemma 3}
\sum_{i=1}^{n}\sqrt{||\textbf{Z}_{i}^{t+1}||^2_2+\epsilon}-\sum_{i=1}^{n}\dfrac{||\textbf{Z}_{i}^{t+1}||^2_2+\epsilon}{2\sqrt{||\textbf{Z}_{i}^{t}||^2_2+\epsilon}} \\ \leq \sum_{i=1}^{n}\sqrt{||\textbf{Z}_{i}^{t}||^2_2+\epsilon}-\sum_{i=1}^{n}\dfrac{||\textbf{Z}_{i}^{t}||^2_2+\epsilon}{2\sqrt{||\textbf{Z}_{i}^{t}||^2_2+\epsilon}}
\end{gathered}
\end{equation}
Subtracting Eq.~(\ref{eq:lemma 3}) from Eq.~(\ref{eq:proof2}), we have 
\begin{equation}
\small
\begin{gathered}
\label{eq:proof4}
\dfrac{1}{2} 
\big(\lVert {\textbf{X} - \textbf{X}\textbf{Z}^{t+1} \rVert}^2_{F} + \alpha \lVert {\textbf{\~X} - \textbf{X}\textbf{\~Z}^{t+1} \rVert}^2_{F}\big) + \lambda_d tr(\textbf{D}^{T}\textbf{Z}^{t+1}) \\
+ \lambda_d tr(\textbf{\~D}^{T}\textbf{\~Z}^{t+1}) 
+ \lambda_s \big(||\textbf{Z}^{t+1}||_{2,1}  +  ||\textbf{\~Z}^{t+1}||_{2,1} \big) + \beta ||{\textbf{Z}_c}^{t+1}||_{2,1} \\ \leq
\dfrac{1}{2} 
\big(\lVert {\textbf{X} - \textbf{X}\textbf{Z}^{t} \rVert}^2_{F} + \alpha \lVert {\textbf{\~X} - \textbf{X}\textbf{\~Z}^{t} \rVert}^2_{F}\big) + \lambda_d tr(\textbf{D}^{T}\textbf{Z}^{t}) \\
+ \lambda_d tr(\textbf{\~D}^{T}\textbf{\~Z}^{t}) 
+ \lambda_s \big(||\textbf{Z}^{t}||_{2,1}  +  ||\textbf{\~Z}^{t}||_{2,1} \big) + \beta ||{\textbf{Z}_c}^{t}||_{2,1}
\end{gathered}
\end{equation}
which establishes that the objective function (\ref{eq:equation5}) monotonically decreases in each iteration. 
Note that the objective function has lower bounds, so it will converge.
Empirical results show that the convergence is fast and only a few iterations are needed to converge.
Therefore, the proposed method can be applied to large scale problems in practice.

\textbf{Acknowledgments} This work was partially supported by NSF grant IIS-1316934. 
We gratefully acknowledge the support of NVIDIA Corporation with the donation of the Tesla K40 GPU used for this research.   

\newpage

{\small
\bibliographystyle{ieee}
\bibliography{egbib}
}

\end{document}